\documentclass[10pt,twocolumn,letterpaper]{article}

\usepackage{multirow}
\usepackage{bbm}
\usepackage{subcaption}
\usepackage{kotex}
\usepackage{xcolor}

\usepackage{wacv}
\usepackage{times}
\usepackage{epsfig}
\usepackage{graphicx}
\usepackage{amsmath}
\usepackage{amssymb}
\usepackage{booktabs}

\newcommand{\subsecspaceA}{\vspace{-.4cm}}

\def\wacvPaperID{1329} 

\wacvfinalcopy 

\ifwacvfinal
\usepackage[breaklinks=true,bookmarks=false]{hyperref}
\else
\usepackage[pagebackref=true,breaklinks=true,colorlinks,bookmarks=false]{hyperref}
\fi

\begin{document}

\title{Domain Adaptive Video Semantic Segmentation via \\ 
Cross-Domain Moving Object Mixing}

\author{Kyusik Cho \quad\quad Suhyeon Lee \quad\quad Hongje Seong \quad\quad Euntai Kim\thanks{}\\
School of Electrical and Electronic Engineering, Yonsei University, Seoul, Korea\\
{\tt\small 
\{ks.cho,hyeon93,hjseong,etkim\}@yonsei.ac.kr}
}


\twocolumn[{
\renewcommand\twocolumn[1][]{#1}
\maketitle
\begin{center}
    \centering
    \includegraphics[width=\linewidth]{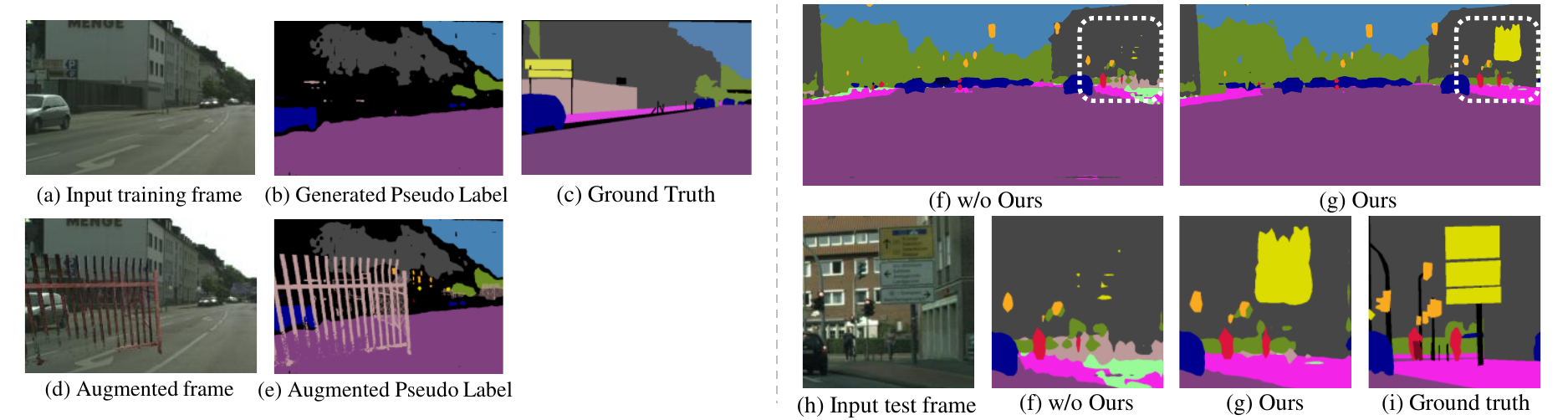}
    \captionof{figure}{
  \textbf{Motivation: self-training with the biased model.} (a) Given the training sample of the target domain, the model biased toward easy-to-transfer classes generates (b) the pseudo-label without hard-to-transfer classes, such as \colorbox[RGB]{244, 35, 232}{\textcolor{white}{\scriptsize \textit{sidewalk}}}, \colorbox[RGB]{190, 153, 153}{\textcolor{black}{\scriptsize \textit{fence}}}, and \colorbox[RGB]{220, 220, 0}{\textcolor{black}{\scriptsize \textit{sign}}}. (c) The ground truth label is shown for visibility. However, the proposed method pastes several classes from the source domain to (d) the target domain video frames and (e) pseudo-labels. (f) The model trained on the \textit{biased} pseudo-labels cannot detect the \colorbox[RGB]{220, 220, 0}{\textcolor{black}{\scriptsize \textit{sign}}} and confused the \colorbox[RGB]{244, 35, 232}{\textcolor{white}{\scriptsize \textit{sidewalk}}} with \colorbox[RGB]{190, 153, 153}{\textcolor{black}{\scriptsize \textit{fence}}} and \colorbox[RGB]{152, 251, 152}{\textcolor{black}{\scriptsize \textit{terrain}}} in the test sample. (g) The model trained with our method detects the classes better and is less noisy in the hard-to-transfer classes.
  } 
    \label{fig_intro}
\end{center}%
}]
{
  \renewcommand{\thefootnote}%
    {\fnsymbol{footnote}}
  \footnotetext[1]{Corresponding author}
}

\begin{abstract}
The network trained for domain adaptation is prone to bias toward the easy-to-transfer classes.
Since the ground truth label on the target domain is unavailable during training, the bias problem leads to skewed predictions, forgetting to predict hard-to-transfer classes.
To address this problem, we propose Cross-domain Moving Object Mixing (CMOM) that cuts several objects, including hard-to-transfer classes, in the source domain video clip and pastes them into the target domain video clip.
Unlike image-level domain adaptation, the temporal context should be maintained to mix moving objects in two different videos.
Therefore, we design CMOM to mix with consecutive video frames, so that unrealistic movements are not occurring.
We additionally propose Feature Alignment with Temporal Context (FATC) to enhance target domain feature discriminability.
FATC exploits the robust source domain features, which are trained with ground truth labels, to learn discriminative target domain features in an unsupervised manner by filtering unreliable predictions with temporal consensus.
We demonstrate the effectiveness of the proposed approaches through extensive experiments. 
In particular, our model reaches mIoU of 53.81\% on VIPER $\rightarrow$ Cityscapes-Seq benchmark and mIoU of 56.31\% on SYNTHIA-Seq $\rightarrow$ Cityscapes-Seq benchmark, surpassing the state-of-the-art methods by large margins. The code is available at: \url{https://github.com/kyusik-cho/CMOM}. 
\end{abstract}

\section{Introduction}
Video semantic segmentation, a task of classifying every pixel in every video frame~\cite{chandra2018deep, jain2019accel, nilsson2018semantic, zhu2017deep}, is a fundamental problem in machine vision. Training the segmentation network requires extensive pixel-level annotated data, which costs expensive human labor. On the other hand, advances in computer technology have made it possible to create perfectly annotated synthetic datasets~\cite{richter2017playing, richter2016playing, ros2016synthia}. However, segmentation networks trained with synthetic data perform poorly in the real world because of the domain gap. 

To handle this issue, many studies of unsupervised domain adaptation (UDA) have been proposed that improve performance in the target domain without any annotations~\cite{hoffman2016fcns, pan2020unsupervised, yang2020fda, zhang2021prototypical}.
These were mainly progressed in image-level, while some advances were recently studied with video data.
DA-VSN~\cite{guan2021domain} has taken a successful first step in the domain adaptive video semantic segmentation (DAVSS) through adversarial training and intra-domain temporal consistency regularization. 
TPS~\cite{TPS_arxiv} applied self-training for video, and Park~\etal~\cite{shin2021unsupervised} used both adversarial training and self-training.
The mainly used approaches, adversarial training and self-training, are well-known solutions in image-level UDA, and they were effectively exploited by extending the promising solutions in image-level to video data. In this paper, we further exploit the self-training.

In UDA, the self-training on target domain is performed by learning with pseudo-labels that are generated via segmentation networks trained on the source domain.
Here, the segmentation networks would generate good quality of the pseudo-labels for easy-to-transfer classes and vice versa.
That means, during self-training, the network will be biased toward easy-to-transfer classes, forgetting to predict hard-to-transfer classes. This problem is depicted in Figure~\ref{fig_intro}.

Inspired by recent studies of UDA for image segmentation~\cite{tranheden2021dacs, wang2020differential}, we design two advanced DAVSS methods to address the bias problem in videos: Cross-domain Moving Object Mixing (CMOM) and Feature Alignment with Temporal Context (FATC).

The first contribution, CMOM, is a data augmentation method that mixes moving objects in the source domain and target domain.
To do this, we can simply adopt the image-level mixing approach~\cite{tranheden2021dacs}.
However, this will break the temporal context of the video, \eg, an object may suddenly disappear in the next frame or a new object may be presented, making it an unrealistic video.
To prevent this problem, CMOM cuts the source domain moving objects from the consecutive video frames and pastes them into the target domain.
The second contribution, FATC, is additionally used to mitigate the bias problem during self-training.
FATC forces to generate discriminative features from target domain videos by reducing distances between robust source features and semantically weak target domain features.
Here, we can also adopt the image-level alignment approach~\cite{wang2020differential}, but this cannot exploit a rich cue of the video: we can predict the results of the current frame not only from the current frame but also from the previous frame by warping the results of the previous frame.
Our FATC effectively exploits this cue by predicting the target domain segmentation mask of the current frame twice and filtering the noisy results.
Then, the remained target domain features are aligned with the source domain features, enabling the network to learn discriminative and robust features in the target domain without accurate ground truth labels.

With two proposed methods, we self-train a semantic segmentation network, and we test on standard DAVSS benchmarks. Our proposal achieves the state-of-the-arts performances on both VIPER~\cite{richter2017playing} $\rightarrow$ Cityscapes-Seq~\cite{cordts2016cityscapes} and SYNTHIA-Seq~\cite{ros2016synthia} $\rightarrow$ Cityscapes-Seq benchmarks, showing the effectiveness of our framework. We also conduct extensive ablation studies to demonstrate that our approach facilitates the networks to learn the robust target domain features in videos.

\begin{figure*}
\begin{center}
\includegraphics[width=\linewidth]{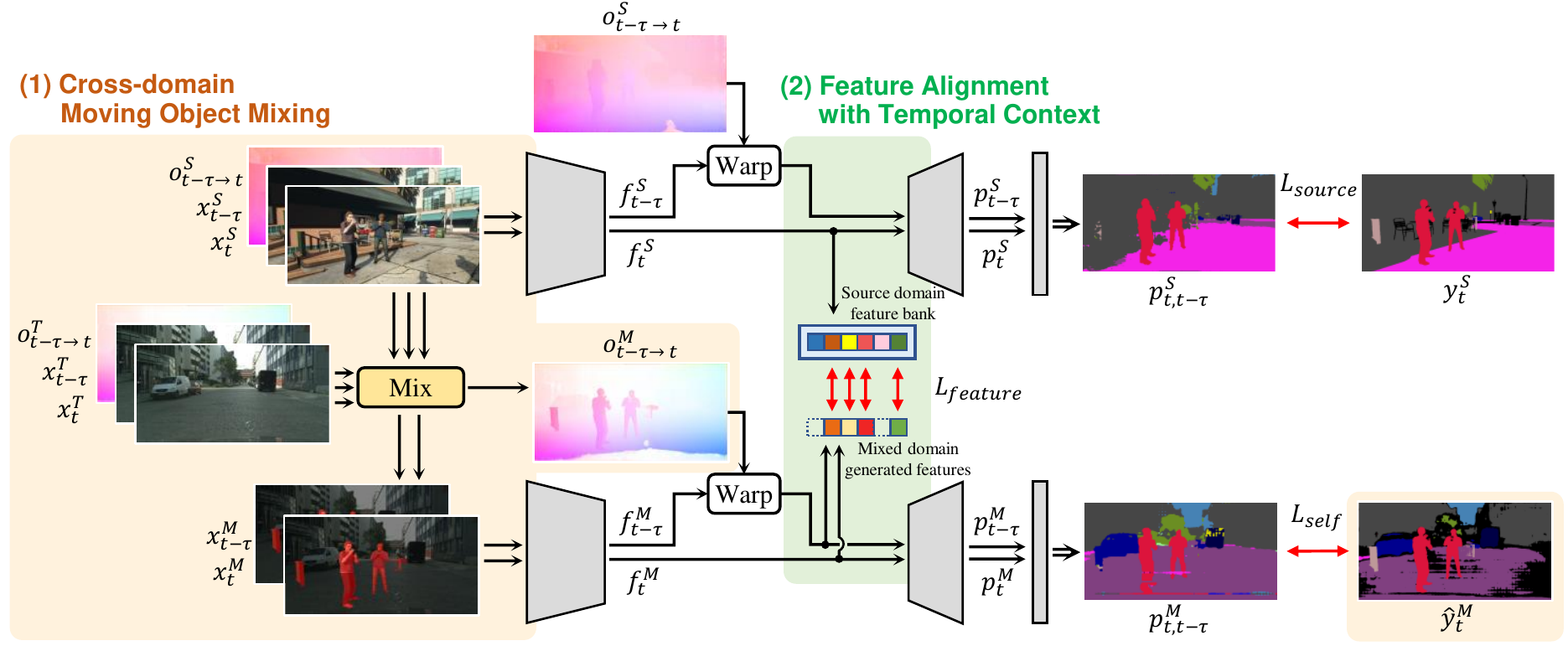}
\end{center}
\vspace{-0.2cm}
   \caption{\textbf{Illustration of the proposed framework.} 
   Our framework is trained with two proposed methods based on self-training. (1) CMOM generates the mixed domain data sample by mixing data samples from two domains. The mixed domain video frames, optical flow, and pseudo-label replace the target domain data sample during the training process. (2) FATC reduces feature mismatch between the generated mixed domain features and source domain features in the feature bank. The unreliable predictions are excluded by comparing the predictions of the previous frame and the current frame. 
   We train the model with three loss terms: source domain segmentation loss $\mathcal{L}_{source}$, mixed domain segmentation loss $\mathcal{L}_{self}$, and feature alignment loss $\mathcal{L}_{feature}$.
}
\vspace{-0.2cm}
\label{fig_framework}
\end{figure*}

\section{Related Work}
\paragraph{Video Semantic Segmentation (VSS).}
VSS is a problem that classifies all pixels in all video frames. The most notable difference between VSS and image semantic segmentation is that VSS can utilize the temporal context of the video.
Two primary goals are considered in utilizing the temporal context, one for accurate predictions and the other one for faster predictions~\cite{wang2021survey}. 
The methods for accurate predictions exploit richer information from multi-frame contexts~\cite{gadde2017semantic, nilsson2018semantic}.
On the other hand, the methods for faster prediction save computation costs by reusing the features extracted from the previous frame~\cite{shelhamer2016clockwork, zhu2017deep}.
Following the previous DAVSS works~\cite{guan2021domain, TPS_arxiv}, we adopt ACCEL~\cite{jain2019accel} as a segmentation network.
ACCEL modularized the network for segmenting previous and current frames to strike a balance between speed and accuracy, making trade-offs a choice.

\paragraph{Domain Adaptive Semantic Segmentation.}
The machine learning algorithm optimizes the model on the training data assuming that the training data distribution and the test data distribution are equal. If this assumption is violated, the model performance drops drastically. The goal of domain adaptation is to minimize this performance drop~\cite{hoffman2016fcns}. 
The most popular methods in unsupervised domain adaptation (UDA) for semantic segmentation are adversarial learning~\cite{guan2021domain, hoffman2016fcns, pan2020unsupervised, Tsai_adaptseg_2018, vu2019advent} and self-training~\cite{huo2022domain,yang2020fda, zhang2021prototypical, zou2018unsupervised}.
Other techniques are also used for domain adaptation: aligning the feature representations of the two domains~\cite{huo2022domain, toldo2021unsupervised, wang2020differential}; augmenting data by changing the source domain image to a target style~\cite{choi2019self, li2019bidirectional, yang2020fda} or mixing two different domain images~\cite{lee2021unsupervised, tranheden2021dacs}; and entropy minimization in the target domain~\cite{vu2019advent}.

\paragraph{Mixing Image Augmentation.}
Data augmentation with mixing images are widely used in various image recognition tasks.
ClassMix~\cite{olsson2021classmix} uses the mixing mask generated based on model prediction to create augmented images and pseudo-labels for unlabeled images, and has proven its effectiveness in semi-supervised semantic segmentation task.
Ghiasi~\etal~\cite{ghiasi2021simple} shows that a simple instance-level mixing only with scale jittering is beneficial for instance segmentation. 
Mixing techniques were also used in UDA~\cite{lee2021unsupervised, tranheden2021dacs} to address the bias problem~\cite{zou2018unsupervised} of self-training approach.
Since the pseudo-label is generated from the model's prediction, the model biased toward easy-to-transfer generates the \textit{biased} pseudo-labels.
The mixing approaches address the bias problem by pasting source domain data into the target domain. 
DACS~\cite{tranheden2021dacs} selects random classes from source images and pastes corresponding pixels to target images. Lee~\etal~\cite{lee2021unsupervised} mix the tail class objects to further solve the class imbalance problem. 
All of the image-level mixings, however, have the drawback that they will break the temporal context of the video by ignoring the movement of objects when they are applied to video. Our approach solves the problem by cutting the source domain moving objects from the consecutive video frames and pasting them into the target domain.

\paragraph{Domain Adaptive Video Semantic Segmentation (DAVSS). }
DAVSS was initially tackled by DA-VSN~\cite{guan2021domain}. DAVSS research mainly focuses on extending image domain adaptation strategies to video using temporal information. For example, DA-VSN~\cite{guan2021domain} uses a sequence of predictions for adversarial learning and enforces consistent prediction in consecutive frames in the target domain. Recently, TPS~\cite{TPS_arxiv} uses consistency learning between frames in the target domain as the main strategy. Meanwhile, Park~\etal~\cite{shin2021unsupervised} conduct DAVSS on different architectures and datasets from them, and use both adversarial learning and self-training. 
Our work also uses self-training, and we further consider the bias problem which has not been explored with video data.

\section{Methodology} 
\subsection{Problem Formulation and Overview}
Let $\mathbb{S}$ and  $\mathbb{T}$ be the source and target domains. 
In the domain adaptive video semantic segmentation (DAVSS) setting, the source domain video frames $\{ x^\mathbb{S}_{1 : T}\}^{N_\mathbb{S}}_{n=1}$ with the corresponding pixel-wise labels $\{y^\mathbb{S}_{1 : T}\}^{N_\mathbb{S}}_{n=1}$ and the unlabeled target domain video frames $\{x^\mathbb{T}_{1 : T}\}^{N_\mathbb{T}}_{n=1}$ are given. 
$T$ is the length of the video, and $N_\mathbb{S}$ and $N_\mathbb{T}$ denotes the number of video samples in the source and target datasets, respectively.
To exploit temporal information, we adopt FlowNet~\cite{ilg2017flownet} to produce the source and target domain optical flows $o^{\mathbb{S}}_{t-\tau \rightarrow t}$ and $o^{\mathbb{T}}_{t-\tau \rightarrow t}$ from time $t-\tau$ to $t$, respectively.
In this paper, we further consider the domain $\mathbb{M}$, a mixture of source and target domains~\cite{tranheden2021dacs}.

The overall framework is illustrated in Figure~\ref{fig_framework}.
Our framework focuses on the self-training based DAVSS with two proposed methods: Cross-domain Moving Object Mixing (CMOM) and Feature Alignment with Temporal Context (FATC). 
At each iteration, source domain video frames $\{ x^\mathbb{S}_{t-\tau}, x^\mathbb{S}_{t}\}$, source domain optical flow $o^{\mathbb{S}}_{t-\tau \rightarrow t}$, mixed domain video frames $\{ x^\mathbb{M}_{t-\tau}, x^\mathbb{M}_{t}\}$, and mixed domain optical flow $o^{\mathbb{M}}_{t-\tau \rightarrow t}$ enter the our video semantic segmentation (VSS) network.
Our network makes source and mixed domain predictions, $ p^\mathbb{S}_{t}$ and $ p^\mathbb{M}_{t}$ at time $t$.
The frame-level predictions of each domain $p^*_{t}$, $p^*_{t-\tau}$ are used to generate a clip-level prediction $p^*_{t, t-\tau}$ with the score fusion module~\cite{jain2019accel}.
For training the network, we provide supervision to both source and mixed domains, $ y^\mathbb{S}_{t}$ and $ \hat{y}^\mathbb{M}_{t}$.
All mixed domain samples are produced by our proposed CMOM.
In the feature space, our FATC excludes noisy features by exploiting the temporal consensus and enhances target domain feature discriminability.

\begin{figure}[t]
\begin{center}
   \includegraphics[width=\linewidth]{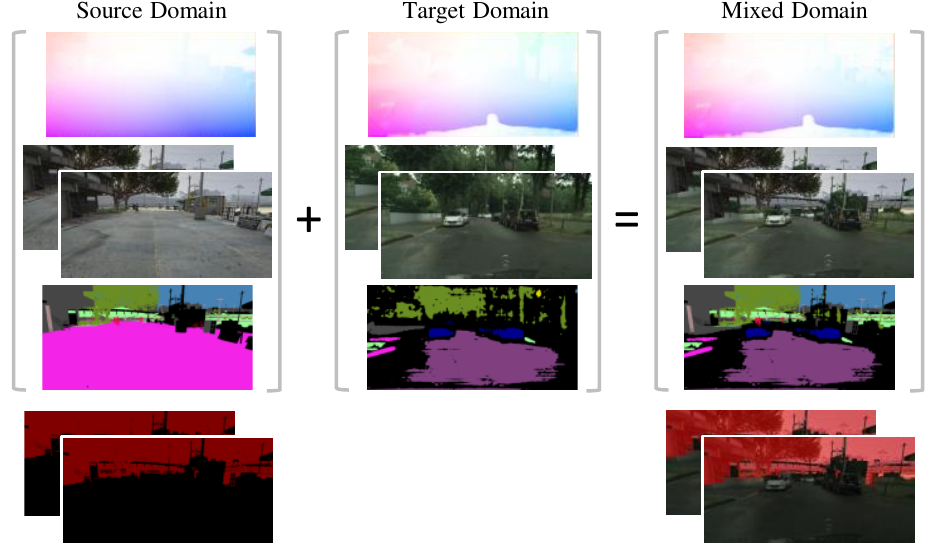}
\end{center}
\vspace{-0.1cm}
   \caption{\textbf{Illustration of CMOM.} CMOM randomly selects some classes in source domain and cuts them from two consecutive frames. Then, they are pasted into target domain. 
   In order to maintain the temporal context of a video, we create a mask with the same classes in all frames constituting the video and mix not only the RGB frame but also the optical flow.
   The last row shows the mixing mask in red for visibility.  
\vspace{-0.1cm}
 }
\label{fig_CMOM}
\end{figure}

\subsection{Cross-Domain Moving Object Mixing}
\label{Section_CMOM}
Our method focuses on the self-training. 
When training the network for DAVSS, due to the absence of the ground truth target labels $\{y^\mathbb{T}_{1 : T}\}^{N_\mathbb{T}}_{n=1}$, the network tends to be biased toward easy-to-transfer classes and forget hard-to-transfer classes.
Therefore, the model hard to generate pseudo-labels with hard-to-transfer classes.
In the domain adaptive image semantic segmentation, DACS~\cite{tranheden2021dacs} alleviates this problem with a cut-and-paste data augmentation strategy between the source and target domain.
DACS data augmentation is formulated as
\begin{equation}
\begin{split} 
\label{DACS}
x^\mathbb{M} = M \odot x^\mathbb{S} + (1-M) \odot x^\mathbb{T},\\
\hat{y}^\mathbb{M} = M \odot y^\mathbb{S} + (1-M) \odot \hat{y}^\mathbb{T},\\
\end{split}
\end{equation}
where $M$ is the pixel selection mask of randomly selected classes, and $\odot$ is the pointwise product. $x^\mathbb{S}$, $x^\mathbb{T}$, and $x^\mathbb{M}$ are the source, target, and mixed domain images, respectively. $y^\mathbb{S}$, $\hat{y}^\mathbb{T}$, and $\hat{y}^\mathbb{M}$ are the source ground truth label, target pseudo-label, and mixed domain pseudo-label, respectively.

Inspired by~\cite{tranheden2021dacs}, we propose a novel data augmentation method CMOM that cuts several objects, including hard-to-transfer classes, in the source domain video clip and pastes them into the target domain video clip. Unlike image-level domain adaptation, the temporal context should be maintained to mix moving objects in two different videos.
Therefore, we design CMOM to mix with consecutive video frames, so that unrealistic movements are not occurring.

The process of CMOM is illustrated in Figure~\ref{fig_CMOM}.
We first randomly select three-quarters of the classes that exist in the current source domain label $y^\mathbb{S}_{t}$.
Secondly, we create binary masks $\{M_{t}, M_{t-\tau}\}$ in which the class pixels selected from the source video labels $\{y^\mathbb{S}_{t}, y^\mathbb{S}_{t-\tau}\}$ are $1$ and the remainders are $0$.
Using the masks, the mixed domain video frames $\{x^\mathbb{M}_{t-\tau}, x^\mathbb{M}_{t}\}$, pseudo-label $\hat{y}^\mathbb{M}_{t}$, and optical flow $o^\mathbb{M}_{t-\tau \rightarrow t}$ are generated by using the following equations:

\begin{equation}
\begin{split}
    x^\mathbb{M}_{t-\tau} &= M_{t-\tau} \odot x^\mathbb{S}_{t-\tau} + (1-M_{t-\tau}) \odot x^\mathbb{T}_{t-\tau}, \\[1pt]
    x^\mathbb{M}_{t} &= M_{t} \odot x^\mathbb{S}_{t} + (1-M_{t}) \odot x^\mathbb{T}_{t}, \\[1pt]
    \hat{y}^\mathbb{M}_{t} &= M_{t} \odot  
    y^\mathbb{S}_{t} + (1-M_{t}) \odot \hat{y}^\mathbb{T}_{t}, \\
    o^\mathbb{M}_{t-\tau \rightarrow t} &= M_{t-\tau} \odot o^\mathbb{S}_{t-\tau \rightarrow t} + (1\!-\!M_{t-\tau}) \odot o^\mathbb{T}_{t-\tau \rightarrow t}.
\end{split}
\label{eq_CMOM}
\end{equation}

In this work, we set the $\tau$ to $1$, and the target pseudo-labels $\{\hat{y}^\mathbb{T}_{1 : T}\}^{N_\mathbb{T}}_{n=1}$ are obtained by using the pre-trained DA-VSN model~\cite{guan2021domain} and IAST pseudo-label policy~\cite{mei2020instance} before the self-training phase. 
This is due to the stability of the network, and they can be replaced by any pre-trained DAVSS model and pseudo-label policy. 

This simple data augmentation method gives us several advantages. First, CMOM provides supervision of all classes, addressing the problem of forgetting to predict hard-to-transfer classes due to bias. In addition, the incorrect pseudo-labels generated by erroneous predictions are alleviated by mixing ground truth labels. It helps to overcome the drawbacks of self-training with biased pseudo-labels. Another advantage is that the temporal context of the video is maintained, which can be easily damaged by the naive image-level approach. To maintain the context, we selected the same object in both frames and mixed the optical flow; the movement of each object is accessible with a sharp optical flow. Lastly, CMOM is data augmentation for videos with low computational cost and simplicity. In order to use image-level adaptation strategies in the video, the method must be applied to all frames. Therefore, approaches that require much computation are burdensome to be used in the video. CMOM does not require much computation and can be used easily. We reported the computational cost in ablation experiments.

CMOM is used as target domain data augmentation in our paper. In other words, the mixed domain video frames $x^\mathbb{M}$ and pseudo-labels $\hat{y}^\mathbb{M}$ are used instead of $x^\mathbb{T}$ and $\hat{y}^\mathbb{T}$ in the training process.

\subsection{Feature Alignment with Temporal Context}
\label{Section_Feature_Align}
Although CMOM helps to enhance the target feature quality by providing supervision of all classes in the mixed domain, the discriminability of the target domain is not as good as that of a fully-supervised trained source domain.
In this section, we propose Feature Alignment with Temporal Context (FATC) for enhancing the discriminability of the target domain.
FATC exploits the robust source domain features, which are trained with ground truth labels, to learn discriminative target domain features in an unsupervised manner by filtering unreliable predictions with temporal consensus. 

Here, we focus on that there are several recent works that focus on the feature alignment for domain adaptive image semantic segmentation~\cite{huo2022domain, toldo2021unsupervised, wang2020differential}, and we can adopt image-level alignment approach to FATC. However, image-level approach cannot exploit a rich cue of the video: we can predict the results of the current frame not only from the current frame but also from the previous frame by warping the results of the previous frame. Focusing on this point, we extend the image-level feature alignment~\cite{wang2020differential} to the DAVSS by performing the feature denoising process. 

\begin{figure}[t]
\begin{center}
   \includegraphics[width=\linewidth]{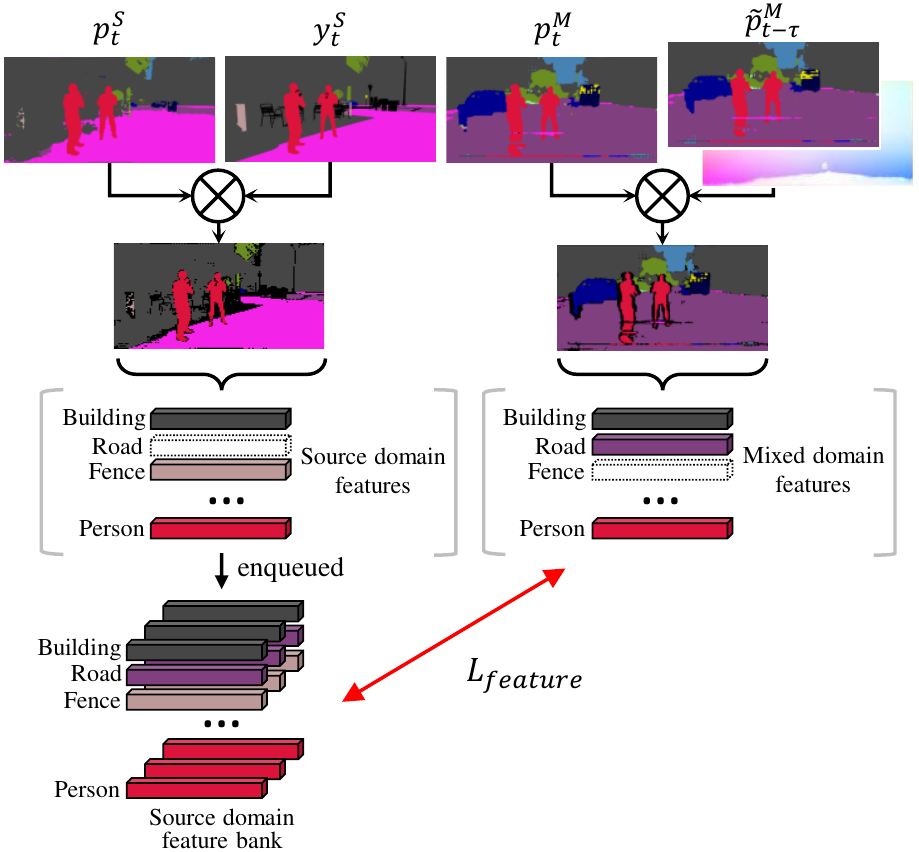}
\end{center}
\vspace{-0.1cm}
   \caption{\textbf{Illustration of FATC.} Feature alignment reduces the mismatch of the features from the two domains. The generated source domain features are stored in the feature bank. The loss function compares each feature in the mixed domain prediction with the closest one in the bank.
   The noisy features are filtered out by comparing with the ground truth label or warped previous frame prediction in the source or mixed domain, respectively.   
    \vspace{-0.2cm}
    }
\label{fig_FeatureAlign}
\end{figure}

Our FATC is illustrated in Figure~\ref{fig_FeatureAlign}. The feature denoising process of FATC excludes noisy features using the temporal context of the video clip. We cannot filter only correct predictions from target domains without ground truth labels, but we can exclude obvious incorrect predictions with video frames. Each video frame is strongly correlated and the corresponding pixel should be predicted to be the same class for the frames. In other words, a consistent prediction between two frames does not guarantee correct answers, but different predictions across frames are always inaccurate. 

We check this temporal correspondence to detect and exclude noisy features.
Using the optical flow $o^{\mathbb{M}}_{t-\tau \rightarrow t}$, we warp $\{p^\mathbb{M}_{t-\tau}, f^\mathbb{M}_{t-\tau}\}$ to $\{\tilde{p}^\mathbb{M}_{t-\tau}, \tilde{f}^\mathbb{M}_{t-\tau}\}$, where $f^{\mathbb{M},c}_t$ indicates the features of each class $c$, $p^\mathbb{M}_t$ is the predictions, and $\tilde{f}^\mathbb{M}_{t-\tau}$ and $\tilde{p}^\mathbb{M}_{t-\tau}$ are the warped features and predictions from time $t-\tau$ to $t$, respectively.
Then, we generate the class-wise valid region binary mask $v^{\mathbb{M},c}_t$ by comparing $p^\mathbb{M}_{t}$ and $\tilde{p}^\mathbb{M}_{t-\tau}$:
\begin{equation}
\begin{split} 
v^{\mathbb{M},c}_t &= \mathbbm{1}^{C}(p^\mathbb{M}_t, \tilde{p}^\mathbb{M}_{t-\tau}), \\[1pt]
V^{\mathbb{M},c}_t &= \{v^{\mathbb{M},c_1}_t, v^{\mathbb{M},c_2}_t, ..., v^{\mathbb{M},c_k}_t\} = \mathcal{T}(v^{\mathbb{M},c}_t).
\end{split}
\end{equation}

Here, $\mathbbm{1}^{C}$ is a class-wise mask generating function that returns $1$ for pixels where $p^{\mathbb{M}}_t = c $ and $p^\mathbb{M}_t = \tilde{p}^\mathbb{M}_{t-\tau}$; and $0$ for otherwise pixels, for $c \in C$. Also, $\mathcal{T}$ is the disconnection detection operator to generate instance-wise features that only works for `things' classes~\cite{wang2020differential}, and $k$ is the number of instance-wise masks in the corresponding class. 
Then, the feature centroid in the mixed domain $\mathcal{F}^{\mathbb{M},c}_t$ is calculated as follows:
\begin{equation}
\begin{split} 
\mathcal{F}^{\mathbb{M},c_k}_t = \frac{\sum v^{\mathbb{M},c_k}_t f^\mathbb{M}_t }{\sum v^{\mathbb{M},c_k}_t}.
\end{split}
\end{equation}

On the other hand, the ground truth labels for the source domain are available. Therefore we exclude features from misclassified pixels using $Y^\mathbb{S}$, as in~\cite{wang2020differential}.
\begin{equation}
\begin{split} 
 v^{\mathbb{S},c}_t &= \mathbbm{1}^{C}(p^\mathbb{S}_t, y^\mathbb{S}_t), \\
 V^{\mathbb{S},c}_t &= \{v^{\mathbb{S},c_1}_t, v^{\mathbb{S},c_2}_t, ..., 
 v^{\mathbb{S},c_k}_t\} = \mathcal{T}(v^{\mathbb{S},c}_t).\\
\end{split}  
\end{equation}
Then, the feature centroid in the source domain $\mathcal{F}^{\mathbb{S},c}_t$ is calculated by the same method as the mixed domain. 
\begin{equation}
 \mathcal{F}^{\mathbb{S},c_k}_t = \frac{\sum v^{\mathbb{S},c_k}_t f^\mathbb{S}_t }{\sum v^{\mathbb{S},c_k}_t}.  
\label{feature_source}  
\end{equation}

Finally, we can adopt image-level feature alignment method~\cite{wang2020differential} to align the denoised source and mixed domain features in the feature space; create a feature bank that stores $\mathcal{F}^{\mathbb{S},c_j}$, and then align $\mathcal{F}^{\mathbb{M},c_i}$ to features in the bank. The feature bank has a fixed size for each class, and the oldest feature is emitted whenever a new $\mathcal{F}^{\mathbb{S},c}$ is generated. The mixed domain foreground features generate the loss with the closest $\mathcal{F}^{\mathbb{S},c_j}$ in the bank, and $i$ and $j$ are the indexes of each feature sample. The loss for feature alignment is L1 loss formulated as:
\begin{equation}
    \mathcal{L}_{feature} = \sum_{i}\min_{j}\frac{\Vert\mathcal{F}^{\mathbb{M},c_i}_t - \mathcal{F}^{\mathbb{S},c_j}_t\Vert_1}{k}.
    \label{L_feature}
\end{equation}

\subsection{Training}
The overall training objective is defined as Eq. (\ref{Loss}), where $\lambda$ controls the balance between the loss terms. $\mathcal{L}_{source}$ and $\mathcal{L}_{self}$ are the supervised loss defined through standard cross entropy loss for the source and mixed domain, respectively.

\begin{equation}
    \mathcal{L} = \mathcal{L}_{source} \; + \; \lambda_{M} \mathcal{L}_{self} + \lambda_{feature}\mathcal{L}_{feature}.
\label{Loss}  
\end{equation}

\renewcommand\arraystretch{1.1}
\begin{table*}[!ht]
\setlength{\tabcolsep}{5pt}
\centering
\begin{small}
\begin{tabular}{p{0.2cm}|p{2cm}|*{2}{p{0.4cm}}|*{15}{p{0.44cm}}|p{0.6cm}}
 \hline \hline
 \multicolumn{20}{c}{\textbf{VIPER~$\rightarrow$~Cityscapes-Seq}} \\
 \hline
 &Methods &Adv. &ST &{road} &{side.} &{buil.} &{fence} &{light} &{sign} &{vege.} &{terr.} &{sky} &{pers.} &{car} &{truck} &{bus} &{mot.} &{bike} &mIoU \\
 \hline 
 \multirow{9}{*}{\rotatebox{90}{Image}}
 &Source only && &56.7 &18.7 &78.7 &6.0 &22.0 &15.6 &81.6 &18.3 &80.4 &59.9 &66.3 &4.5 &16.8 &20.4 &10.3 &37.1 \\
 &AdvEnt~\cite{vu2019advent} &\multicolumn{1}{c}{\checkmark}& &78.5 &31.0 &81.5 &22.1 &29.2 &26.6 &81.8 &13.7 &80.5 &58.3 &64.0 &6.9 &38.4 &4.6 &1.3 &41.2 \\
 &CBST~\cite{zou2018unsupervised} &&\multicolumn{1}{c|}{\checkmark} &48.1 &20.2 &84.8 &12.0 &20.6 &19.2 &83.8 &18.4 &\textbf{84.9} &59.2 &71.5 &3.2 &38.0 &23.8 &37.7 &41.7 \\
 &IDA~\cite{pan2020unsupervised} &\multicolumn{1}{c}{\checkmark}& &78.7 &33.9 &82.3 &22.7 &28.5 &26.7 &82.5 &15.6 &79.7 &58.1 &64.2 &6.4 &41.2 &6.2 &3.1 &42.0 \\
 &CRST~\cite{zou2019confidence} &&\multicolumn{1}{c|}{\checkmark} &56.0 &23.1 &82.1 &11.6 &18.7 &17.2 &85.5 &17.5 &82.3 &60.8 &73.6 &3.6 &38.9 &\textbf{30.5} &35.0 &42.4 \\
 &SVMin~\cite{guan2021scale} &\multicolumn{1}{c}{\checkmark}&\multicolumn{1}{c|}{\checkmark} &51.1 &14.3 &80.8 &11.9 &30.9 &23.1 &83.5 &\textbf{37.7} &74.5 &59.5 &\textbf{79.7} &36.4 &\textbf{53.2} &20.0 &4.2 &44.1    \\
 &CrCDA~\cite{huang2020contextual} &\multicolumn{1}{c}{\checkmark}& &78.1 &33.3 &82.2 &21.3 &29.1 &26.8 &82.9 &28.5 &80.7 &59.0 &73.8 &16.5 &41.4 &7.8 &2.5 &44.3\\
 &RDA~\cite{huang2021rda} &&\multicolumn{1}{c|}{\checkmark} &72.0 &25.9 &80.8 &15.1 &27.2 &20.3 &82.6 &31.4 &82.2 &56.3 &75.5 &22.8 &48.3 &19.1 &6.7 &44.4\\
 &FDA~\cite{yang2020fda} &&\multicolumn{1}{c|}{\checkmark} &70.3 &27.7 &81.3 &17.6 &25.8 &20.0 &83.7 &31.3 &82.9 &57.1 &72.2 &22.4 &49.0 &17.2 &7.5 &44.4 \\ \hline
 \multirow{3}{*}{\rotatebox{90}{Video}}
 &DA-VSN~\cite{guan2021domain} &\multicolumn{1}{c}{\checkmark}& &86.8 &36.7 &83.5 &22.9 &30.2 &27.7 &83.6 &26.7 &80.3 &60.0 &79.1 &20.3 &47.2 &21.2 &11.4 &47.8\\
 &TPS~\cite{TPS_arxiv} &&\multicolumn{1}{c|}{\checkmark} &82.4 &36.9 &79.5 &9.0 &26.3 &29.4 &78.5 &28.2 &81.8 &61.2 &80.2 &\textbf{39.8} &40.3 &28.5 &31.7 &48.9\\
 &\textbf{Ours} &&~\checkmark &\textbf{89.0} &\textbf{53.8} &\textbf{86.8} &\textbf{31.0} &\textbf{32.5} &\textbf{47.3} &\textbf{85.6} &25.1 &80.4 &\textbf{65.1} &79.3 &21.6 &43.4 &25.7 &\textbf{40.6} &\textbf{53.8}\\
 \hline \hline
\end{tabular}
\end{small}
\vspace{0.1cm}
\caption{ \textbf{Quantitative comparisons on VIPER~$\rightarrow$~Cityscapes-Seq.} Results are expressed in per-class IoU and mIoU scores. Our method achieved the highest IoU in most classes, and a significant update to the mIoU compared with the state-of-the-art methods. \textit{Image:} the methods originally proposed for domain adaptive image semantic segmentation. These studies were re-experimented in a structure for video segmentation. \textit{Video:} the methods proposed for domain adaptive video semantic segmentation. \textit{Adv.:} Domain adaptation methods based on adversarial training. \textit{ST:} self-training.}
\label{table_viper}
\end{table*}

\vspace{-0.15cm}

\begin{table*}[!ht]
\centering
\begin{small}
\begin{tabular}{p{0.2cm}|p{2cm}|*{2}{p{0.4cm}}|*{11}{p{0.6cm}}|p{0.6cm}}
 \hline \hline
 \multicolumn{16}{c}{\textbf{SYNTHIA-Seq~$\rightarrow$~Cityscapes-Seq}} \\
 \hline
 &Methods &Adv. &ST &{road} &{side.} &{buil.} &{pole} &{light} &{sign} &{vege.} &{sky} &{pers.} &{rider} &{car} &mIoU \\
 \hline
 \multirow{9}{*}{\rotatebox{90}{Image}}
 &Source only && &56.3 &26.6 &75.6 &25.5 &5.7 &15.6 &71.0 &58.5 &41.7 &17.1 &27.9 &38.3 \\
  &AdvEnt~\cite{vu2019advent} &\multicolumn{1}{c}{\checkmark} & &85.7 &21.3 &70.9 &21.8 &4.8 &15.3 &59.5 &62.4 &46.8 &16.3 &64.6 &42.7 \\
  &CBST~\cite{zou2018unsupervised} &&\multicolumn{1}{c|}{\checkmark} &64.1 &30.5 &78.2 &28.9 &14.3 &21.3 &75.8 &62.6 &46.9 &20.2 &33.9 &43.3 \\
  &IDA~\cite{pan2020unsupervised} &\multicolumn{1}{c}{\checkmark}& &87.0 &23.2 &71.3 &22.1 &4.1 &14.9 &58.8 &67.5 &45.2 &17.0 &73.4 &44.0 \\
  &CRST~\cite{zou2019confidence} &&\multicolumn{1}{c|}{\checkmark} &70.4 &31.4 &79.1 &27.6 &11.5 &20.7 &78.0 &67.2 &49.5 &17.1 &39.6 &44.7 \\
  &SVMin~\cite{guan2021scale} &\multicolumn{1}{c}{\checkmark}&\multicolumn{1}{c|}{\checkmark} &84.9 &0.5 &77.9 &29.6 &7.4 &15.0 &78.6 &73.2 &46.9 &6.2 &73.8 &44.9 \\
  &CrCDA~\cite{huang2020contextual} &\multicolumn{1}{c}{\checkmark} &&86.5	&26.3	&74.8	&24.5	&5.0	&15.5	&63.5	&64.4	&46.0	&15.8	&72.8	&45.0\\
  &RDA~\cite{huang2021rda} &&\multicolumn{1}{c|}{\checkmark} &84.7	&26.4	&73.9	&23.8	&7.1	&18.6	&66.7	&68.0	&48.6	&9.3	&68.8	&45.1\\
  &FDA~\cite{yang2020fda} &&\multicolumn{1}{c|}{\checkmark} &84.1 &32.8 &67.6 &28.1 &5.5 &20.3 &61.1 &64.8 &43.1 &19.0 &70.6 &45.2 \\
\hline 
\multirow{3}{*}{\rotatebox{90}{Video}}
 &DA-VSN~\cite{guan2021domain} &\multicolumn{1}{c}{\checkmark}&&89.4 &31.0 &77.4 &26.1 &9.1 &20.4 &75.4 &74.6 &42.9 &16.1 &82.4 &49.5 \\
 &TPS~\cite{TPS_arxiv} &&\multicolumn{1}{c|}{\checkmark}&\textbf{91.2} &\textbf{53.7} &74.9 &24.6 &\textbf{17.9} &\textbf{39.3} &68.1 &59.7 &57.2 &\textbf{20.3} &\textbf{84.5} &53.8 \\
 &\textbf{Ours} &&~\checkmark &90.4 &39.2 &\textbf{82.3} &\textbf{30.2} &16.3 &29.6 &\textbf{83.2} &\textbf{84.9} &\textbf{59.3} &19.7 &84.3 &\textbf{56.3}\\
\hline \hline
\end{tabular}
\end{small}
\vspace{0.1cm}
\caption{\textbf{Quantitative comparisons on SYNTHIA-Seq~$\rightarrow$~Cityscapes-Seq.}
}
\label{table_syn}
\end{table*}

\begin{figure*}
\begin{center}
\includegraphics[width=0.9\linewidth]{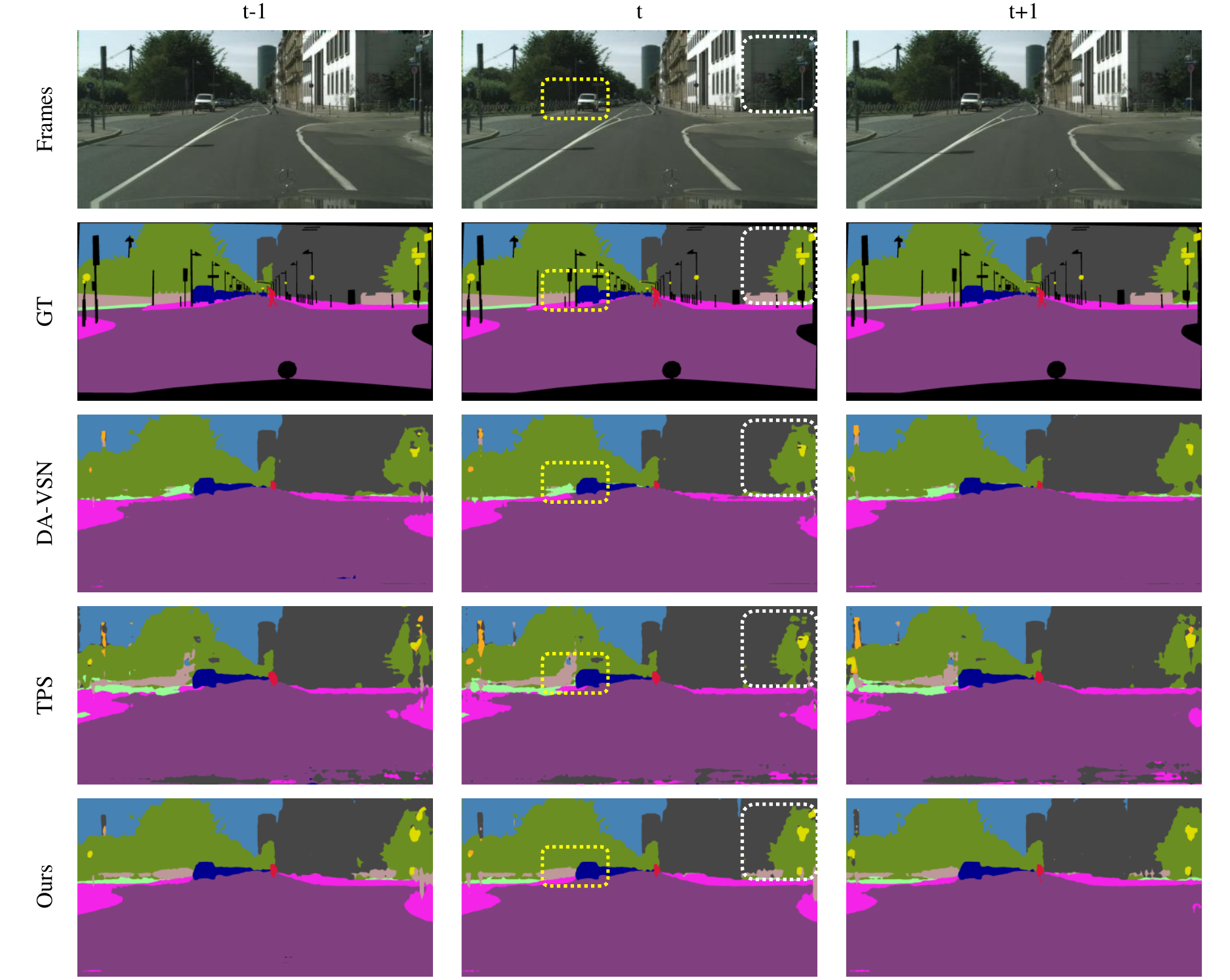}
\end{center}
\vspace{-0.15cm}
   \caption{\textbf{Qualitative result.} We compare our qualitative results with other state-of-the-art methods. The models are trained over VIPER $\rightarrow$ Cityscapes-Seq benchmark. Since the ground truth of Cityscapes-Seq is provided only one frame per 30 frames, we show the same ground truth at time $t$ for all frames. 
   }
\vspace{-0.15cm}
\label{fig_qualitative}
\end{figure*}

\section{Experiments}
\subsection{Setup}
\paragraph{Architecture.} Following the previous works for domain adaptive video segmentation~\cite{guan2021domain, TPS_arxiv}, we use ACCEL~\cite{jain2019accel} architecture for video semantic segmentation. The architecture consists of two segmentation branches, an optical flow estimation module, and a score fusion module. Each segmentation branch is designed with ResNet-101~\cite{he2016deep} backbone and DeepLab~\cite{chen2017deeplab} classification head to obtain single-frame prediction. The optical flow estimation module and score fusion module are used to combine each single-frame level prediction into video-level prediction. FlowNet~\cite{ilg2017flownet} is used as the optical flow estimation module, and the score fusion module is a single $1 \times 1$ convolutional layer.

\subsecspaceA
\paragraph{Datasets.} 
Following \cite{guan2021domain} and \cite{TPS_arxiv}, we use VIPER~\cite{richter2017playing}, SYNTHIA-Seq~\cite{ros2016synthia}, and Cityscapes-Seq~\cite{cordts2016cityscapes} datasets for benchmarking.
Cityscapes-Seq is a benchmark set consisting of 5000 video clips captured from real streets. This set is split into 2975, 500, and 1525 clips for training, validation, and testing, respectively. Each video clip consists of 30 frames, and the ground truth label is provided only in a single frame (20th frame). VIPER is a synthetic dataset generated with the game `Grand Theft Auto V'. VIPER contains 134K video frames and corresponding segmentation labels. SYNTHIA-Seq is another synthetic dataset containing 8000 frames and corresponding segmentation labels.

\subsecspaceA
\paragraph{Evaluation Protocols.} 
Following the previous DAVSS methods~\cite{guan2021domain, TPS_arxiv}, we measure the Intersection over Union (IoU) score on the validation set of Cityscapes-Seq. IoU score is measured for common classes of the source and target domain for evaluation. 15 common classes are selected for VIPER to Cityscapes-Seq task, and 11 common classes are selected for SYNTHIA-Seq to Cityscapes-Seq task.

\subsecspaceA
\paragraph{Training Details.} To train our network, We use an SGD optimizer with a momentum of 0.9 and weight decay of $5\times10^{-4}$. The initial learning rate was set to $5\times10^{-4}$ and decreased according to the polynomial decay with a power of 0.9. We trained our network for 40K iterations. The balancing parameters in Eq. (\ref{Loss}) are set to $\lambda_M = 1$, $\lambda_{feature} = 0.01$. 
The feature bank size is set to 50.
Our model is initialized with the pretrained model of DA-VSN~\cite{guan2021domain} and then we start to self-train.
For self-training, we generate pseudo-labels offline with our proposed methods and IAST~\cite{mei2020instance} policy that adjusts the threshold of the prediction score for each class and each image.
The pseudo-label hyperparameters of IAST~\cite{mei2020instance} are set to $\alpha=0.2$, $\beta = 0.9$, $\gamma=8$.

\begin{table*}
\centering
\begin{subtable}{0.5\linewidth}
{
\centering
\setlength{\tabcolsep}{9.5pt}
\resizebox{0.95\columnwidth}{!}
{
\begin{small}
\begin{tabular}{c|ccc|c}
 \hline \hline
 \multicolumn{5}{c}{\textbf{VIPER~$\rightarrow$~Cityscapes-Seq}} \\
 \hline
 Method &{ST} &{CMOM} &{FATC} &mIoU \\
 \hline
  Baseline & & & &47.85 \\
  Only Self-Training &\checkmark & & &49.89 \\
  with CMOM &\checkmark &\checkmark & &53.72 \\
  with FATC &\checkmark & &\checkmark &50.86 \\
  Ours &\checkmark &\checkmark &\checkmark &\textbf{53.81} \\
\hline \hline
\end{tabular}
\end{small}
}
\vspace{-0.1cm}
\caption{Ablation study on loss functions. \textit{ST:} self-training}
\label{subtable_ablation_loss}
\vspace{0.2cm}

\setlength{\tabcolsep}{18pt}
\resizebox{0.95\columnwidth}{!}
{
\begin{small}
\begin{tabular}{c|cc}
 \hline \hline
 \multicolumn{3}{c}{\textbf{Learning time}} \\
 \hline
 Method &w/o CMOM &w/ CMOM \\
  Time (s/iter) &2.031 &2.087\\
\hline \hline
\end{tabular}
\end{small}
}
\vspace{-0.1cm}
\caption{Ablation study on learning time with CMOM.}
\label{subtable_ablation_learningtime}
}

\end{subtable}%
\begin{subtable}{0.5\linewidth}
{
\centering
\setlength{\tabcolsep}{5pt}
\resizebox{0.95\columnwidth}{!}
{
\begin{small}
\begin{tabular}{c|cc|cc|c}
 \hline \hline
 \multicolumn{6}{c}{\textbf{VIPER~$\rightarrow$~Cityscapes-Seq}} \\
 \hline
 Mix Class &Things &Stuffs  &Movable &Stationary  &All\\
 mIoU &52.07 &52.86  &52.54 &53.28  &\textbf{53.81} \\ 
\hline \hline
\end{tabular}
\end{small}
}
\vspace{-0.15cm}
\caption{ CMOM mixing contents. }
\label{subtable_ablation_mix_things}

\setlength{\tabcolsep}{8pt}
\resizebox{0.95\columnwidth}{!}
{
\begin{small}
\begin{tabular}{c|ccccc}
 \hline \hline
 \multicolumn{6}{c}{\textbf{VIPER~$\rightarrow$~Cityscapes-Seq}} \\
 \hline
 Mix ratio  &0\% &25\% &50\% &75\% &100\% \\
 mIoU  & 50.86 & 53.12 & 53.59 &\textbf{53.81} &{40.98} \\ 
\hline \hline
\end{tabular}
\end{small}
}
\vspace{-0.15cm}
\caption{CMOM mixing ratio.}
\label{subtable_ablation_mix_ratio}

\setlength{\tabcolsep}{7.5pt}
\resizebox{0.95\columnwidth}{!}
{
\begin{small}
\renewcommand{\arraystretch}{1.0}
\begin{tabular}{c|ccc|ccc}
 \hline \hline 
 \multicolumn{1}{c|}{\renewcommand{\arraystretch}{0.85}\begin{tabular}[c]{@{}c@{}} ~ \\ ~ \end{tabular}} 
 &\multicolumn{3}{c|}{\renewcommand{\arraystretch}{0.6}\begin{tabular}[c]{@{}c@{}}\textbf{VIPER~$\rightarrow$} \\ 
 \textbf{Cityscapes-Seq}\end{tabular}}  &\multicolumn{3}{c}{\renewcommand{\arraystretch}{0.6}\begin{tabular}[c]{@{}c@{}}\textbf{SYNTHIA-Seq~$\rightarrow$} \\ \textbf{Cityscapes-Seq}\end{tabular}} \\
 
 \hline
\multicolumn{1}{c|}{Method} & \multicolumn{1}{c}{TPS} & \multicolumn{1}{c}{+Ours} & \multicolumn{1}{c|}{Gain} & \multicolumn{1}{c}{TPS} & \multicolumn{1}{c}{+Ours} & \multicolumn{1}{c}{Gain} \\
\multicolumn{1}{c|}{mIoU}   & \multicolumn{1}{c}{48.9} & \multicolumn{1}{c}{53.9} & \multicolumn{1}{c|}{+5.0} & \multicolumn{1}{c}{53.8} & \multicolumn{1}{c}{58.8} & \multicolumn{1}{c}{+5.0} \\
\hline \hline
\end{tabular}
\end{small}
}
\vspace{-0.15cm}
\caption{ Complementary study on TPS~\cite{TPS_arxiv}. }
\label{subtable_ablation_TPS}
}
\end{subtable}
\caption{\textbf{Ablation Experiments.}
\vspace{-0.5cm}
}
\vspace{-0.2cm}
\end{table*}

\subsection{Comparison with State-of-the-art}
In Tables~\ref{table_viper} and~\ref{table_syn}, we compare our framework with the state-of-the-art methods on VIPER $\rightarrow$ Cityscapes-Seq and SYNTHIA-Seq $\rightarrow$ Cityscapes-Seq benchmarks. To our best knowledge, DA-VSN~\cite{guan2021domain} and TPS~\cite{TPS_arxiv} are the only works, published after peer review, that study domain adaptive video semantic segmentation on the same architectures and datasets as ours. We compare the performance of the proposed method with these baselines and with multiple domain adaptive image segmentation baselines~\cite{guan2021scale, huang2021rda, huang2020contextual, pan2020unsupervised, vu2019advent, yang2020fda, zou2019confidence, zou2018unsupervised}. These baselines are based on various strategies: adversarial training~\cite{guan2021scale, huang2020contextual, pan2020unsupervised, vu2019advent}, self-training~\cite{guan2021scale, huang2021rda, yang2020fda, zou2019confidence, zou2018unsupervised}, and data augmentation~\cite{huang2021rda, yang2020fda}.
All methods were tested on the same architecture for video segmentation. 
As shown in Table~\ref{table_viper}, our framework, the model trained with CMOM and FATC, achieved mIoU of 53.81\% and outperforms all baselines by a large margin. In addition, as shown in Table~\ref{table_syn}, we achieve the best performance with mIoU of 56.31\%, surpassing all baselines on SYNTHIA-Seq $\rightarrow$ Cityscapes-Seq scenario. Our work significantly outperforms state-of-the-art methods in both scenarios, demonstrating the effectiveness of our approach.

In addition, we present qualitative comparisons with the state-of-the-art methods in Figure~\ref{fig_qualitative}.
As shown in the qualitative results, DA-VSN~\cite{guan2021domain} and TPS~\cite{TPS_arxiv} cannot accurately predict hard-to-transfer classes, \ie, signs (denoted in yellow boxes) and fences (denoted in white boxes), while our approach shows good results.
Furthermore, we can observe that our results seem less noisy than the other methods.

\subsection{Ablation Experiments}
We present several ablation experiments to demonstrate the effectiveness of our method. In this section, we tested on VIPER $\rightarrow$ Cityscapes-Seq benchmark unless mentioned otherwise.

\subsecspaceA
\paragraph{Computational costs of CMOM.}
To show the computational cost of CMOM, we measure the training time with and without CMOM. 
The results are summarized in Table~\ref{subtable_ablation_learningtime}. 
Compared to ``without CMOM'', only 2.8\% ($= 2.087/2.031 - 1$) of learning time is additionally required for training with CMOM. 
Therefore, CMOM does not require much computation but it is effective in training.

\subsecspaceA
\paragraph{Loss Functions.}
We ablate the proposed loss functions, and the results are given in Table~\ref{subtable_ablation_loss}. As shown in the table, naively training with simple pseudo-labels achieves a performance improvement marginally (+2.04\%). With either CMOM or FATC, we surpass the simple self-training approach. Furthermore, we achieve significant performance improvement from baseline with both CMOM and FATC by 5.96\%.

\subsecspaceA
\paragraph{Mixing Ratio.}
In Table~\ref{subtable_ablation_mix_ratio}, we present experimental results with various mixing ratios.
We conduct the experiments with mixing ratios of 0\%, 25\%, 50\%, 75\%, and 100\%.
Note that the mixed data at mixing ratios of 0\% and 100\% are the same as target and source data, respectively. As shown in Table~\ref{subtable_ablation_mix_ratio}, we empirically find that the mixing ratio of 75\% achieves the best performance. In addition, the experimental results of mixing ratio 25\%, 50\%, and 75\% are significantly improved compared to performance without mixing and achieved similar performance. This means that our approach is valid regardless of the mixing ratio.

\subsecspaceA
\paragraph{Mixing Contents.}
Things (\eg, person, sign) and stuff (\eg, road, sidewalk) are well-known categories in panoptic segmentation~\cite{kirillov2019panoptic}. 
However, in the context of \textit{moving object mixing}, several classes are movable (\eg, person, car) while others are stationary (\eg, light, sign). 
In this study, we perform CMOM with things, stuff, movable, or stationary classes. The results are given in Table~\ref{subtable_ablation_mix_things}.
Interestingly, we find that mixing with all classes achieves the best performance.
We conjuncture that every class has movement in the video clip because the data has been collected from moving camera, so every class can be considered movable.

\subsecspaceA
\paragraph{Complementary Study.} Table~\ref{subtable_ablation_TPS} shows the performance of our framework built on TPS~\cite{TPS_arxiv}. As a result of the experiment, our approach boosted the performance of TPS by +5.0 mIoU in both benchmarks. Experimental results show that our approach is not limited to a specific framework and is well generalized.

\section{Conclusion}
In this paper, we introduce a novel DAVSS framework. The proposed CMOM and FATC effectively address the bias problem which is caused by self-training in videos.
We demonstrate the effectiveness of the proposed methods with comparison experiments and extensive ablation studies on VIPER $\rightarrow$ Cityscapes-Seq and SYNTHIA-Seq $\rightarrow$ Cityscapes-Seq benchmarks.
We believe that our motivation and insight in this paper can be an essential step to addressing the problem caused by self-training in videos.

\subsecspaceA
\paragraph{Acknowledgements.} This work was supported by the KIST Institutional Program (Project No. 2E31571-22-143).

{\small
\bibliographystyle{ieee_fullname}
\bibliography{main.bbl}

\begin{thebibliography}{10}\itemsep=-1pt

\bibitem{chandra2018deep}
Siddhartha Chandra, Camille Couprie, and Iasonas Kokkinos.
\newblock Deep spatio-temporal random fields for efficient video segmentation.
\newblock In {\em Proceedings of the IEEE Conference on Computer Vision and
  Pattern Recognition}, pages 8915--8924, 2018.

\bibitem{chen2017deeplab}
Liang-Chieh Chen, George Papandreou, Iasonas Kokkinos, Kevin Murphy, and Alan~L
  Yuille.
\newblock Deeplab: Semantic image segmentation with deep convolutional nets,
  atrous convolution, and fully connected crfs.
\newblock {\em IEEE transactions on pattern analysis and machine intelligence},
  40(4):834--848, 2017.

\bibitem{choi2019self}
Jaehoon Choi, Taekyung Kim, and Changick Kim.
\newblock Self-ensembling with gan-based data augmentation for domain
  adaptation in semantic segmentation.
\newblock In {\em Proceedings of the IEEE/CVF International Conference on
  Computer Vision}, pages 6830--6840, 2019.

\bibitem{cordts2016cityscapes}
Marius Cordts, Mohamed Omran, Sebastian Ramos, Timo Rehfeld, Markus Enzweiler,
  Rodrigo Benenson, Uwe Franke, Stefan Roth, and Bernt Schiele.
\newblock The cityscapes dataset for semantic urban scene understanding.
\newblock In {\em Proceedings of the IEEE Conference on Computer Vision and
  Pattern Recognition}, pages 3213--3223, 2016.

\bibitem{gadde2017semantic}
Raghudeep Gadde, Varun Jampani, and Peter~V Gehler.
\newblock Semantic video cnns through representation warping.
\newblock In {\em Proceedings of the IEEE International Conference on Computer
  Vision}, pages 4453--4462, 2017.

\bibitem{ghiasi2021simple}
Golnaz Ghiasi, Yin Cui, Aravind Srinivas, Rui Qian, Tsung-Yi Lin, Ekin~D Cubuk,
  Quoc~V Le, and Barret Zoph.
\newblock Simple copy-paste is a strong data augmentation method for instance
  segmentation.
\newblock In {\em Proceedings of the IEEE/CVF Conference on Computer Vision and
  Pattern Recognition}, pages 2918--2928, 2021.

\bibitem{guan2021scale}
Dayan Guan, Jiaxing Huang, Shijian Lu, and Aoran Xiao.
\newblock Scale variance minimization for unsupervised domain adaptation in
  image segmentation.
\newblock {\em Pattern Recognition}, 112:107764, 2021.

\bibitem{guan2021domain}
Dayan Guan, Jiaxing Huang, Aoran Xiao, and Shijian Lu.
\newblock Domain adaptive video segmentation via temporal consistency
  regularization.
\newblock In {\em Proceedings of the IEEE International Conference on Computer
  Vision}, pages 8053--8064, 2021.

\bibitem{he2016deep}
Kaiming He, Xiangyu Zhang, Shaoqing Ren, and Jian Sun.
\newblock Deep residual learning for image recognition.
\newblock In {\em Proceedings of the IEEE Conference on Computer Vision and
  Pattern Recognition}, pages 770--778, 2016.

\bibitem{hoffman2016fcns}
Judy Hoffman, Dequan Wang, Fisher Yu, and Trevor Darrell.
\newblock Fcns in the wild: Pixel-level adversarial and constraint-based
  adaptation.
\newblock {\em arXiv preprint arXiv:1612.02649}, 2016.

\bibitem{huang2021rda}
Jiaxing Huang, Dayan Guan, Aoran Xiao, and Shijian Lu.
\newblock Rda: Robust domain adaptation via fourier adversarial attacking.
\newblock {\em arXiv preprint arXiv:2106.02874}, 2021.

\bibitem{huang2020contextual}
Jiaxing Huang, Shijian Lu, Dayan Guan, and Xiaobing Zhang.
\newblock Contextual-relation consistent domain adaptation for semantic
  segmentation.
\newblock In {\em The European Conference on Computer Vision}, pages 705--722.
  Springer, 2020.

\bibitem{huo2022domain}
Xinyue Huo, Lingxi Xie, Hengtong Hu, Wengang Zhou, Houqiang Li, and Qi Tian.
\newblock Domain-agnostic prior for transfer semantic segmentation.
\newblock In {\em Proceedings of the IEEE/CVF Conference on Computer Vision and
  Pattern Recognition}, pages 7075--7085, 2022.

\bibitem{ilg2017flownet}
Eddy Ilg, Nikolaus Mayer, Tonmoy Saikia, Margret Keuper, Alexey Dosovitskiy,
  and Thomas Brox.
\newblock Flownet 2.0: Evolution of optical flow estimation with deep networks.
\newblock In {\em Proceedings of the IEEE Conference on Computer Vision and
  Pattern Recognition}, pages 2462--2470, 2017.

\bibitem{jain2019accel}
Samvit Jain, Xin Wang, and Joseph~E Gonzalez.
\newblock Accel: A corrective fusion network for efficient semantic
  segmentation on video.
\newblock In {\em Proceedings of the IEEE Conference on Computer Vision and
  Pattern Recognition}, pages 8866--8875, 2019.

\bibitem{kirillov2019panoptic}
Alexander Kirillov, Kaiming He, Ross Girshick, Carsten Rother, and Piotr
  Doll{\'a}r.
\newblock Panoptic segmentation.
\newblock In {\em Proceedings of the IEEE/CVF Conference on Computer Vision and
  Pattern Recognition}, pages 9404--9413, 2019.

\bibitem{lee2021unsupervised}
Suhyeon Lee, Junhyuk Hyun, Hongje Seong, and Euntai Kim.
\newblock Unsupervised domain adaptation for semantic segmentation by content
  transfer.
\newblock In {\em Proceedings of the AAAI Conference on Artificial
  Intelligence}, volume~35, pages 8306--8315, 2021.

\bibitem{li2019bidirectional}
Yunsheng Li, Lu Yuan, and Nuno Vasconcelos.
\newblock Bidirectional learning for domain adaptation of semantic
  segmentation.
\newblock In {\em Proceedings of the IEEE/CVF Conference on Computer Vision and
  Pattern Recognition}, pages 6936--6945, 2019.

\bibitem{mei2020instance}
Ke Mei, Chuang Zhu, Jiaqi Zou, and Shanghang Zhang.
\newblock Instance adaptive self-training for unsupervised domain adaptation.
\newblock In {\em The European Conference on Computer Vision}, pages 415--430.
  Springer, 2020.

\bibitem{nilsson2018semantic}
David Nilsson and Cristian Sminchisescu.
\newblock Semantic video segmentation by gated recurrent flow propagation.
\newblock In {\em Proceedings of the IEEE conference on computer vision and
  pattern recognition}, pages 6819--6828, 2018.

\bibitem{olsson2021classmix}
Viktor Olsson, Wilhelm Tranheden, Juliano Pinto, and Lennart Svensson.
\newblock Classmix: Segmentation-based data augmentation for semi-supervised
  learning.
\newblock In {\em Proceedings of the IEEE/CVF Winter Conference on Applications
  of Computer Vision}, pages 1369--1378, 2021.

\bibitem{pan2020unsupervised}
Fei Pan, Inkyu Shin, Francois Rameau, Seokju Lee, and In~So Kweon.
\newblock Unsupervised intra-domain adaptation for semantic segmentation
  through self-supervision.
\newblock {\em arXiv preprint arXiv:2004.07703}, 2020.

\bibitem{richter2017playing}
Stephan~R Richter, Zeeshan Hayder, and Vladlen Koltun.
\newblock Playing for benchmarks.
\newblock In {\em Proceedings of the IEEE International Conference on Computer
  Vision}, pages 2213--2222, 2017.

\bibitem{richter2016playing}
Stephan~R Richter, Vibhav Vineet, Stefan Roth, and Vladlen Koltun.
\newblock Playing for data: Ground truth from computer games.
\newblock In {\em European conference on computer vision}, pages 102--118.
  Springer, 2016.

\bibitem{ros2016synthia}
German Ros, Laura Sellart, Joanna Materzynska, David Vazquez, and Antonio~M
  Lopez.
\newblock The synthia dataset: A large collection of synthetic images for
  semantic segmentation of urban scenes.
\newblock In {\em Proceedings of the IEEE Conference on Computer Vision and
  Pattern Recognition}, pages 3234--3243, 2016.

\bibitem{shelhamer2016clockwork}
Evan Shelhamer, Kate Rakelly, Judy Hoffman, and Trevor Darrell.
\newblock Clockwork convnets for video semantic segmentation.
\newblock In {\em European Conference on Computer Vision}, pages 852--868.
  Springer, 2016.

\bibitem{shin2021unsupervised}
Inkyu Shin, Kwanyong Park, Sanghyun Woo, and In~So Kweon.
\newblock Unsupervised domain adaptation for video semantic segmentation.
\newblock {\em arXiv preprint arXiv:2107.11052}, 2021.

\bibitem{toldo2021unsupervised}
Marco Toldo, Umberto Michieli, and Pietro Zanuttigh.
\newblock Unsupervised domain adaptation in semantic segmentation via
  orthogonal and clustered embeddings.
\newblock In {\em Proceedings of the IEEE/CVF Winter Conference on Applications
  of Computer Vision}, pages 1358--1368, 2021.

\bibitem{tranheden2021dacs}
Wilhelm Tranheden, Viktor Olsson, Juliano Pinto, and Lennart Svensson.
\newblock Dacs: Domain adaptation via cross-domain mixed sampling.
\newblock In {\em Proceedings of the IEEE Winter Conference on Applications of
  Computer Vision}, pages 1379--1389, 2021.

\bibitem{Tsai_adaptseg_2018}
Y.-H. Tsai, W.-C. Hung, S. Schulter, K. Sohn, M.-H. Yang, and M. Chandraker.
\newblock Learning to adapt structured output space for semantic segmentation.
\newblock In {\em IEEE Conference on Computer Vision and Pattern Recognition
  (CVPR)}, 2018.

\bibitem{vu2019advent}
Tuan-Hung Vu, Himalaya Jain, Maxime Bucher, Matthieu Cord, and Patrick
  P{\'e}rez.
\newblock Advent: Adversarial entropy minimization for domain adaptation in
  semantic segmentation.
\newblock In {\em Proceedings of the IEEE Conference on Computer Vision and
  Pattern Recognition}, pages 2517--2526, 2019.

\bibitem{wang2021survey}
Wenguan Wang, Tianfei Zhou, Fatih Porikli, David Crandall, and Luc Van~Gool.
\newblock A survey on deep learning technique for video segmentation.
\newblock {\em arXiv preprint arXiv:2107.01153}, 2021.

\bibitem{wang2020differential}
Zhonghao Wang, Mo Yu, Yunchao Wei, Rogerio Feris, Jinjun Xiong, Wen-mei Hwu,
  Thomas~S Huang, and Honghui Shi.
\newblock Differential treatment for stuff and things: A simple unsupervised
  domain adaptation method for semantic segmentation.
\newblock In {\em Proceedings of the IEEE Conference on Computer Vision and
  Pattern Recognition}, pages 12635--12644, 2020.

\bibitem{TPS_arxiv}
Yun Xing, Dayan Guan, Jiaxing Huang, and Shijian Lu.
\newblock Domain adaptive video segmentation via temporal pseudo supervision.
\newblock In {\em European Conference on Computer Vision}, 2022.

\bibitem{yang2020fda}
Yanchao Yang and Stefano Soatto.
\newblock Fda: Fourier domain adaptation for semantic segmentation.
\newblock In {\em Proceedings of the IEEE Conference on Computer Vision and
  Pattern Recognition}, pages 4085--4095, 2020.

\bibitem{zhang2021prototypical}
Pan Zhang, Bo Zhang, Ting Zhang, Dong Chen, Yong Wang, and Fang Wen.
\newblock Prototypical pseudo label denoising and target structure learning for
  domain adaptive semantic segmentation.
\newblock In {\em Proceedings of the IEEE/CVF conference on computer vision and
  pattern recognition}, pages 12414--12424, 2021.

\bibitem{zhu2017deep}
Xizhou Zhu, Yuwen Xiong, Jifeng Dai, Lu Yuan, and Yichen Wei.
\newblock Deep feature flow for video recognition.
\newblock In {\em Proceedings of the IEEE conference on computer vision and
  pattern recognition}, pages 2349--2358, 2017.

\bibitem{zou2019confidence}
Yang Zou, Zhiding Yu, Xiaofeng Liu, BVK Kumar, and Jinsong Wang.
\newblock Confidence regularized self-training.
\newblock In {\em Proceedings of the IEEE International Conference on Computer
  Vision}, pages 5982--5991, 2019.

\bibitem{zou2018unsupervised}
Yang Zou, Zhiding Yu, BVK Vijaya~Kumar, and Jinsong Wang.
\newblock Unsupervised domain adaptation for semantic segmentation via
  class-balanced self-training.
\newblock In {\em Proceedings of the European Conference on Computer Vision},
  pages 289--305, 2018.

\end{thebibliography}
}

\clearpage

\appendix

\section{Visualization of FATC }
\begin{figure}[h]
\begin{center}
   \includegraphics[width=\linewidth]{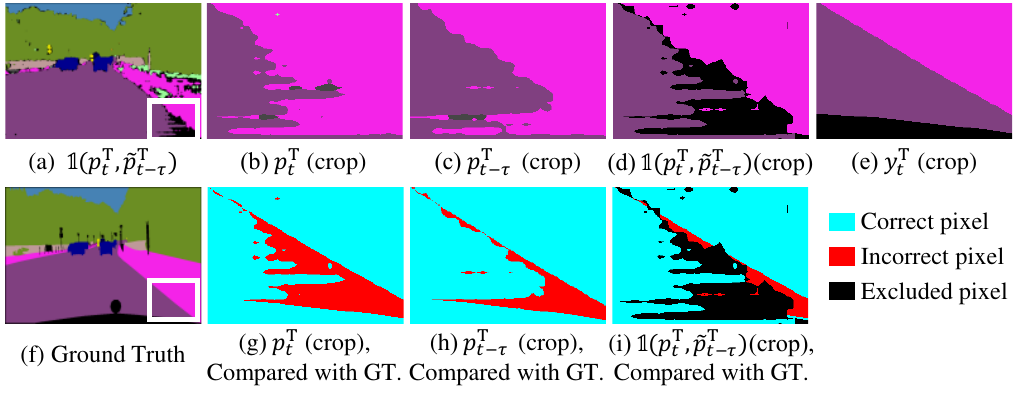}
\end{center}
\caption{\textbf{Effect of FATC.} The first row shows semantic segmentation maps and the second row shows error maps. 
In the error map, correct pixels are colored in blue, wrong predicted pixels are colored in red, and pixels excluded by temporal consensus are colored in black. 
}
\label{supp_FATC}
\end{figure}

We propose Feature Alignment with Temporal Context (FATC) to filter out unreliable predictions with temporal consensus. To show the robustness of the filtering idea, we visualize an example of the process in Figure~\ref{supp_FATC}. 
For (a) the prediction, we zoom a certain region and show (b) the current frame prediction ($p^\mathbb{T}_t$), (c) previous frame prediction ($p^\mathbb{T}_{t-\tau}$), (d) result after filtering with temporal context ($\mathbbm{1}(p^\mathbb{T}_{t},p^\mathbb{T}_{t-\tau})$), and (e) ground truth ($y^\mathbb{T}_t$). (f) Corresponding ground truth for the full image is shown for visibility.
In the second row, we additionally show (g-i) the error maps. Here, blue-colored pixels indicate the correct prediction, the reds indicate the wrong predicted pixels, and the blacks indicate the removed pixels. 
As shown in the figure, the filtering with temporal context removes the error regions in the current frame ($p^\mathbb{T}_t$) effectively. 
Since our feature alignment algorithm generated the feature centroid with blue and red colored pixels, our filtering algorithm will contribute to make more robust feature centroid.

\section{Reproducibility}
To show the stability and reproducibility of our methods, we present additional results of our method by training the network three times with different random seeds. We obtain the results by mIoU of 53.81, 53.53, and 54.78 (average 54.04 $\pm$ 0.54). The three experimental results show that our proposal is stable and reproducible.

\section{More Quantitative Results}
We further present additional qualitative results for real-world videos from Cityscapes demoVideo~\cite{cordts2016cityscapes} dataset. The results are available online: \url{https://youtu.be/xrfe21mNQh0}

\end{document}